\pdfoutput=1

\documentclass{article}
\usepackage{url} 
\usepackage{array} 
\usepackage{amsthm, amsmath, graphicx, amsfonts, mathrsfs, float, enumerate, multirow, rotating}
\usepackage{longtable, subcaption}
\usepackage{array}
\usepackage{multicol,multirow}
\usepackage[labelfont=bf]{caption}
\usepackage{hyperref}

\usepackage{subcaption}
\DeclareCaptionSubType*[Alph]{table}
\DeclareCaptionLabelFormat{mystyle}{Table~\bothIfFirst{#1}{ ̃}#2}
\usepackage[a4paper]{geometry}
\newcolumntype{M}[1]{>{\centering\arraybackslash}m{#1}}
\newcolumntype{N}{@{}m{0pt}@{}}

\begin{document}
\title{Review of state-of-the-arts in artificial intelligence\\
with application to AI safety problem}
\author{Vladimir Shakirov  \\ 
\it{MIPT, Moscow} \\
schw90@mail.ru\\ \\ \\
\date{}
}

\maketitle
\thispagestyle{empty}
\pagestyle{plain}

\begin{abstract}
Here, I review current state-of-the-arts in many areas of AI to estimate when it's reasonable to expect human level AI development. Predictions of prominent AI researchers vary broadly from very pessimistic predictions of Andrew Ng to much more moderate predictions of Geoffrey Hinton and optimistic predictions of Shane Legg, DeepMind cofounder. Given huge rate of progress in recent years and this broad range of predictions of AI experts, AI safety questions are also discussed. 
\end{abstract}
\setcounter{page}{1}

\section{Introduction}
Recent progress in deep learning algorithms for artificial intelligence has raised widespread ethical concerns~\cite{AIasaPositiveAndNegative}\cite{FoomDebate}. It has been argued that human-level AI isn't automatically good for humanity. It might be presumptuous and overconfident to be sure that humans would be able to control superhuman-clever AIs, that those AIs would really care about humans, for example to allow us full access to mineral resources and agriculture fields of the planet. While there are numerous advantages of having clever AIs in the short-term, the long-term danger of having too clever AIs might outweigh, leading to net negative effect of AI progress on society. The most common argument against consideration of such long-term risks is their vagueness due to supposed very long time distance from us~\cite{andrewNgpredictions2}. 

The main purpose of my article is to show that superhuman level AI is reasonably possible even within next 5 to 10 years. Certainly it's very hard to predict future of science. However it's still a battle of arguments: whatever side brings more convincing arguments wins even if arguments from both sides are not very convincing leading to low margin of win (like 60\%/40\% instead of 99\%/1\%). In this article the recent progress in deep learning is reviewed. This progress has some regularities which allow to extrapolate it into the future. Such extrapolation speaks in favor of reasonably high probability of human level AI in next 5 to 10 years. So "long-term risks" are quite probably not that distant and vague. 

The most similar article has been written in 2013 by Katja Grace~\cite{GraceProgress}. Another review with more details on deep history of deep learning has been written in 2014 by J{\"u}rgen Schmidhuber~\cite{1404.7828}.

The structure of the article is as follows. In chapters 2 - 5, I review some state-of-the-arts in natural language processing, computer vision, speech recognition and reinforcement learning. In chapters 6 - 8, I answer some questions which often arise in scientific discussions about how much time we have until human level AI: unsupervised and multimodal learning, the need for embodiment. In chapters 9 and 10, I review some relevant details of the progress in neuroscience. In chapter 11, some new and prospective approaches are reviewed. Many of them might revolutionize modern deep learning but we don't know yet which ones. In chapter 12, the predictions of prominent scientists about when human level AI would be developed are discussed. In chapter 13 and 14, I make a brief review of AI safety research.

\section{Natural language processing}

Let's begin with some state-of-the-arts in natural language generation.\\

\begin{table}[H]
\caption{Best perplexity scores}
\label{nce-vs-sampled}
\vskip 0.15in
\begin{center}
\begin{small}
$
\begin{tabular}{|lcc|}
\hline
dataset & perplexity & link \\

\hline
Wikipedia english corpus snapshot 2014/09/17  (1.5B words) & 27.1 & ~\cite{1602.06291} \\
\hline
1B word benchmark (shuffled sentences) & 24.2 & ~\cite{1602.02410}  \\
\hline
OpenSubtitles (923M words) & 17 & ~\cite{1506.05869} \\
\hline
IT Helpdesk Troubleshooting (30M words) & 8 & ~\cite{1506.05869} \\
\hline
Movie Triplets (1M words) & 27 & ~\cite{1507.04808} \\
\hline
PTB (1M words) & 62.34 & ~\cite{1511.03962} \\
\hline
\end{tabular}
$
\end{small}
\end{center}
\vskip -0.1in
\end{table}

Why perplexity matters?\\
If neural net says smth inconsistent (in \textbf{any} sense: logical, syntactic, pragmatic) then it means that it gives too much probability to some inappropriate words i.e, it's perplexity isn't optimized yet. When hierarchical neural chat bots would achieve low enough perplexity, they would likely to write coherent stories, to answer intelligibly with common sense, to reason in a consistent and logical way etc. Just as in \cite{1508.06576} we would be able to adjust a conversational model to imitate style and opinions of a distinct person.\\

\textbf{What are reasonable predictions for perplexity improvements in the nearest future?}\\ 
Let's take two impressive recent works both submitted to arxiv in February 2016. They are: "contextual LSTM models for large scale NLP tasks"~\cite{1602.06291} and "Exploring the limits of language modeling"~\cite{1602.02410}.\\
\begin{table}[H]
\caption{Best perplexity scores, \textbf{single} model}
\label{nce-vs-sampled}
\vskip 0.15in
\begin{center}
\begin{small}
$
\begin{tabular}{|lcc|}
\hline
number of hidden neurons & perplexity in ~\cite{1602.06291} & perplexity in ~\cite{1602.02410}\\
\hline
256 & 38 & -- \\
\hline
512 & -- & 54 \\
\hline
1024 & 27 & --  \\
\hline
2048 & -- & 44 \\
\hline
4096 & -- & -- \\
\hline
8192 & -- & 30 \\
& & 24.2 (ensemble) \\
\hline
\end{tabular}
$
\end{small}
\end{center}
\vskip -0.1in
\end{table}

From table 2, it's quite reasonable to predict that for \cite{1602.06291} hs=4096 might give pplx$<$20, hs=8192 might give pplx$<$15 and ensemble of models with hs=8192 trained on 10B words might give \textbf{perplexity well below 10.} Nobody can tell now what kind of common sense reasoning would such a neural net have.\\

It's quite probable that even in this year the discussed decrease in perplexity might allow us to create neural chat bots that write reasonable stories, answer intelligibly with common sense, discuss things. All this would be just a mere consequence of low enough perplexity. \\

Shannon estimated lower and upper bounds of human perplexity to be 0.6 and 1.3 bits per character \cite{shannon1950}. Strictly speaking, applying formula (17) in his work gives lower bound equal 0.648 which he rounded. Given average word length of 4.5 symbols and including spaces (as Shannon included them in his game) gives us \textbf{an estimate for lower bound on human-level word perplexity as 11.8} = $2^{0.648*5.5}$ or better to say 10 plus something. This lower bound might be much less than real human perplexity \cite{mmahoney}. Another source of improvement may come from solving some discrepancy in what kind of perplexity is optimized \cite{1511.05101} \cite{1511.01844}. However, both KL(P$||$Q) and KL(Q$||$P) have optimum when P=Q. Also, \cite{1511.05101} \cite{1511.01844} propose ways to partially solve that problem using adversarial learning. "Generating sequences from continuous space"\cite{1511.06349} demonstrates impressive advantages of adversarial paradigm.\\

\begin{table}[H]
\caption{Best BLEU scores}
\label{nce-vs-sampled}
\vskip 0.15in
\begin{center}
\begin{small}
$
\begin{tabular}{|lccc|}
\hline
lang pair & BLEU & dataset & link \\
\hline
en$=>$fr & 37.5 & WMT'14 & ~\cite{1410.8206}     \\
\hline
fr$=>$en & 35.8 & WMT'14 & ~\cite{1601.00372}\\
\hline
ar$=>$en & 56.4 & NIST OpenMT'12 & ~\cite{1506.00698}  \\
\hline
ch$=>$en & 40.06 & MT03 & ~\cite{1512.02433} \\
\hline
ge$=>$en & 29.3 & WMT'15 & ~\cite{1511.06709}  \\
\hline
en$=>$ge & 26.5 & WMT'15 & ~\cite{1511.06709}  \\
\hline
en$=>$ru & 29.37 & newstest-14 & ~\cite{1603.06147} \\
\hline
en$=>$ja & 36.21 & WAT'15 & ~\cite{ZhongyuanZhu} \\

\hline
\end{tabular}
$
\end{small}
\end{center}
\vskip -0.1in
\end{table}

Human BLEU score for chinese=$>$english translation on MT03 dataset is  35.76 \cite{humanchenbleu}. 
In recent article \cite{1512.02433} neural network gets 40.06 BLEU on the same task and dataset.
They took state-of-the-art \cite{1409.0473} "GroundHog" network and replaced maximum likelihood estimation with their own MRT criterion, which increased BLEU from 33.2 to 40.06. Here is a quote from abstract: "Unlike conventional maximum likelihood estimation, minimum risk training is capable of optimizing model parameters directly with respect to evaluation metrics". Another impressive improvement comes from improving translation with monolingual data~\cite{1511.06709}. Modern neural nets translate text $\sim$1000 times faster than humans do~\cite{devlin}. Study of foreign languages is now of less practical use because NMT improves faster than most humans can.\\

\section{Computer vision}
\begin{table}[H]
\caption{Performance improvement on several tasks}
\label{nce-vs-sampled}
\vskip 0.15in
\begin{center}
\begin{small}
$
\begin{tabular}{|lcccc|}
\hline
year & ImageNet top-5 error & ImageNet localization & PASCAL VOC detection & SPORTS-1M video classification\\
\hline
2011 & 25,77\% & 42,5\% & & \\
\hline
2012 & 15,31\% & 33,5\% & & \\
\hline
2013 & 11,20\% & 29,9\% & $<$50\% & \\
\hline
2014 & 6,66\% & 25,3\% & 63,8\% & 63.9\% \\
\hline
2015 & 3,57\% & 9,0\% & 76,4\% & 73.1\% \\
\hline
\end{tabular}
$
\end{small}
\end{center}
\vskip -0.1in
\end{table}

"Identity mappings in deep residual networks"\cite{1603.05027} reach 5.3\% top-5 error in single  one-crop model while human level is reported to be 5.1\%\cite{KarpathyWhatIlearnedfrom}. In "deep residual networks" \cite{1512.03385} one-crop single model gives 6.7\% but ensemble of those models with Inception gives 3.08\%\cite{1602.07261} Just to notice that $6.7 : 3.08 = 5.3 : 2.4$. Another big improvement comes from "deep networks with stochastic depth"\cite{1603.09382}. There was reported $\sim0.3$\% error in human annotations to ImageNet\cite{KarpathyWhatIlearnedfrom}, so \textbf{real error on ImageNet would soon become even below 2\%.} Human level is overcome not only in ImageNet classification task (see also an efficient implementation of 21841 classes ImageNet classification\cite{21481clustersImageNet}) but also on boundary detection\cite{1511.07386}. Video classification task on SPORTS-1M dataset (487 classes, 1M videos) performance improved from 63.9\%\cite{karpathySPORTS1M} (2014) to 73.1\%\cite{1503.08909} (March 2015). See also \cite{2015breakthroughYearCV}.\\ 

CNNs outperform humans also in terms of speed being \textbf{$\sim$1000x faster than human} \cite{soumithBenchmarks} (note that times there are given for batches) or even $\sim$10 000x times faster after compression \cite{1602.01528}. Video processing 24fps on AlexNet demands just 82 Gflops and GoogleNet demands 265 Gflops. Here's why. The best benchmark in\cite{soumithBenchmarks} gives 25ms feedforward and 71ms total time for 128 pictures batch on NVIDIA Titan X (6144 Gflops), so for 24fps video real-time feedforward processing we need 6144 Gflops * 24/128 * 0.025 = 30 Gflops. If we want to learn something using backprop than we need 6144 Gflops * 24/128 * 0.071 = 82 Gflops. Same calculations for GoogleNet give 83 Gflops and 265 Gflops respectively.\\

Nets answer questions based on images \cite{1603.01417}. Using similar method as in \cite{1505.00468} the equivalent human age of net~\cite{1603.01417} can be estimated as 6.2 years old (submitted 4 Mar 2016), while \cite{1511.02274} was 5.45 years (submitted 7 Nov 2015), \cite{1505.00468} was 4.45 years (submitted 3 May 2015). See appendix 1 for details of age estimates.  Also, nets can describe images with sentences, in some metrics even better than humans can~\cite{COCOleaderboard}. Beside video=$>$text \cite{NeuralTalkAnimator}\cite{1511.03476}\cite{1510.07712}, there are some experiments to realize text=$>$picture \cite{1603.01801}\cite{1512.00570}\cite{1511.02793}.

\section{Speech recognition and generation}

\begin{table}[H]
\caption{speech recognition performance errors}
\label{nce-vs-sampled}
\vskip 0.15in
\begin{center}
\begin{small}
$
\begin{tabular}{|lccccc|}
\hline
year & CHiME noisy & VoxForge European & WSJ eval'93 & LibriSpeech test-other & citation \\
\hline
2014 & 67.94\% & 31.2\% & 6.94\% & 21.74\% & \cite{1412.5567} \\
\hline
2015 & 21.79\% & 17.55\% & 4.98\% & 13.25\% & \cite{1512.02595} \\
\hline
human & 11.84\% & 12.76\% & 8.08\% & 12.69\% & \cite{1512.02595} \\
\hline
\end{tabular}
$
\end{small}
\end{center}
\vskip -0.1in
\end{table}

Here, AI hasn't surpassed human level yet but it's clearly seen that we have all chances to see it in 2016. The rate of improvement is very fast. For example, Google reports it's word error rate dropped from 23\% in 2013 to 8\% in 2015 \cite{googleWERdrop}. \\

\section{Reinforcement learning}
AlphaGo and Atari agents choose one action from a finite set of possible actions. There are also articles on \textit{continuous} reinforcement learning where action is a vector \cite{1603.00748} \cite{1506.02438} \cite{1509.02971} \cite{1603.00448} \cite{1511.04143} \cite{1510.09142} \cite{1512.04455} \cite{1509.03005} \cite{1512.07679}.\\

AlphaGo is a powerful AGI in it's own very simple world which is just a board with stones and a bunch of simple rules. If we improve AlphaGo with continuous reinforcement learning (and if we manage to make it work well on hard real-world tasks) than it would be real AGI in a real world. Also, we can begin with pretraining it on virtual videogames\cite{togelius}. Videogames often contain even much more interesting challenges than real life gives for an average human. Millions of available books and videos contain the concentrated life experience of millions of people. While AlphaGo is a successful example of pairing modern RL with modern CNN, RL can be combined with neural chat bots and reasoners \cite{1511.04636}\cite{1506.08941}.  \\


\section{do we have powerful unsupervised learning?}
Yes, we do. DCGAN \cite{1511.06434} generates reasonable pictures\cite{LeCunDCGANpost}.
Language generation models which minimize perplexity are unsupervised and were improved very much recently (table 1). "Skip-thought vectors"\cite{1506.06726} generate vector representation for sentences allowing to train \textit{linear} classifiers over those vectors and their cosine distances to solve many supervised problems at state-of-the-art level. Generative nets gained deep new horizons at the edge of 2013-14 \cite{1312.6114}\cite{1401.4082}. Recent work \cite{1603.08575} continues progress in "computer vision as inverse graphics" approach. \\

\section{Is embodiment critical?}
People with tetra-amelia syndrome~\cite{tetraamelia} have neither hands nor legs from their birth and they still manage to get a good intelligence. For example, Hirotada Ototake~\cite{HirotadaOtokake} is a Japanese sports writer famous for his bestseller memoirs. He also worked as a school teacher. Nick Vujicic~\cite{NickVujicic} has written many books, graduated from Griffith University with a Bachelor of Commerce degree, he often reads motivational lectures. Prince Randian~\cite{princeRandian} spoke Hindi, English, French, and German. Modern robots have better embodiment.\\

There are quick and good drones, there are quite impressive results in robotic grasping~\cite{googlegrasp}. It's argued that embodiment might be done in virtual world of videogames\cite{togelius}. Among 8 to 18-year-olds average amount of time spent with TV/computer/video games etc is 7.5 hours a day~\cite{childreninscreens}. According to another study~\cite{adultsinscreens} UK adults spend an average an 8 hours and 40 minutes a day on media devices. Humans develop commonsense intelligence very fast. When you are 2 years old you barely can do something that current AI can't. When you're 4 years old you have common sense, you can learn from texts and conversations. However, 2 years are just 100 weeks. If we exclude sleep we get $\sim$50 weeks. The 24fps video input of 50 weeks might be processed in 10 hours on pretrained AlexNet using four Titan X.

\section{do we have multimodal learning?}
Yes, we do. Multimodal learning is used in \cite{1509.06086}\cite{1412.6632}\cite{1503.01800}\cite{1412.3121} to improve video classification related tasks. Unsupervised multimodal learning is used in grounding of textual phrases in images \cite{1511.03745} using attention-based mechanism so that different modalities supervise one another. In "neural self-talk" \cite{1512.03460} a neural network sees a picture, generates questions based on it and answers those questions itself.

\section{Argument from neuroscience}

This is a simplified view of human cortex:
\begin{figure}[H]
\begin{subfigure}[b]{1\textwidth}
\centering
\includegraphics[scale=0.5, trim=0 0mm 0mm 0, clip]{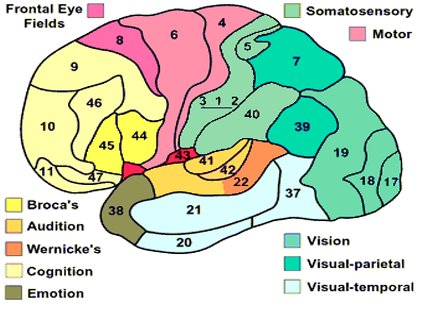}\\
\label{ClusterImg_group}
\end{subfigure}
\caption{Human brain, lateral view.}
\label{ClusterImg}
\end{figure} 
Roughly speaking~\cite{chudlerfacts}~\cite{kennedycc}, 15\% of human brain is devoted to low-level vision tasks (occipital lobe).\\
Another 15\% are devoted to image and action recognition (somewhat more than a half of temporal lobe).\\
Another 15\% are devoted to objects detection and tracking (parietal lobe).\\
Another 15\% are devoted to speech recognition and pronunciation (BAs 41,42,22,39,44, parts of 6,4,21).\\
Another 10\% are devoted to reinforcement learning (orbitofrontal cortex and part of medial prefrontal cortex).
Together they represent about 70\% of human brain.\\

\textbf{Modern neural networks work at about human level for these 70\% of human brain.}\\
For example, CNNs make 1.5x less mistakes than humans at ImageNet while acting about 1000x times faster than a human. \\

\subsection{Remaining 30\% of human brain cortex}
From the neuroscience view, human cortex has the similar structure throughout all its surface.\\
It's just ~3mm thick mash of neurons functioning on the same principles throughout all the cortex.\\
There is likely no big difference between how prefrontal cortex works and how other parts of cortex work.\\ 
There is likely no big difference in their speed of calculations, in complexity of their algorithms.\\
It would be somewhat strange if modern deep neural networks can't solve remaining 30\% in several years.\\

About 10\% are low-level motorics (BAs 6,8). Robotic hands are not very dexterous. However people having no fingers from their birth have problems with fine motorics but still develop normal intelligence ~\cite{tetraamelia}, see also the section "Is embodiment critical?". Also, one of DLPFC functions is attention which is actively used now in LSTMs.\\

The only part which still lacks near human-level performance are BAs 9,10,46,45 which constitute together only 20\% of human brain cortex. These areas are responsible for complex reasoning, complex tools usage, complex language. However, "A neural conversational model"\cite{1506.05869} , "Contextual LSTM..."\cite{1602.06291}, "playing Atari with deep reinforcement learning"\cite{1312.5602}, "mastering the game of Go..."\cite{AlphaGoNature} and numerous other already mentioned articles have recently begun to really attack this problem. \\

There seem to be no particular reasons to expect these 30\% to be much harder than other 70\%. Less than 3 years has gone from AlexNet winning ImageNet competition to surpassing human level. It's reasonable to expect the same $<$ 3 years gap from "A neural conversational model" winning over Cleverbot to human-level reasoning. After all, there are much more deep learning researchers now with much more knowledge and experience, there are much more companies interested in DL.

\section{Do we know how our brain works?}
Detailed connectome deciphering has begun to really succeed recently with creation of multi-beam scanning electron microscopy \cite{firstMultibeamSEM} (Jan 2015), which allowed labs to get a grant for deciphering the detailed connectome of 1x1x1 mm$^3$ of rat cortex~\cite{1x1x1dream} after the proof-of-principle deciphering of 40x40x50 mcm of brain cortex with 3x3x30 nm resolution was achieved~\cite{40x40x50}(July 2015).\\

Works \cite{1510.05067}\cite{1411.0247} show that weight symmetry is not important for backpropagation in many tasks: one may propagate errors using a fixed matrix with everything still working. This counterintuitive conclusion is a crucial step towards theoretical understanding of how brain manages to work. See also "Towards Biologically Plausible Deep Learning" by Y.Bengio et.al~\cite{1502.04156}. There arise some theoretic elements of how brain might perform credit assignment in deep hierarchies \cite{1510.02777}.\\ 

Recently, STDP objective function has been proposed\cite{1509.05936}. It's like an unsupervised objective function somewhat similar to what is used for example in word2vec. In \cite{1506.06472} authors surveyed a space of polynomial local learning rules (learning rules in brain are supposed to be local) and found that backpropagation outperforms them. There are also online learning approaches which require no backpropagation through time, f.e.\cite{1507.07680}. Though they can't compete with conventional deep learning, brain perhaps can use something like that given fantastic number of it's neurons and synaptic connections.

\section{So what separates us from human-level AI?}
Recently, a flow of articles about memory networks and neural turing machines made it possible to use arbitrarily large memories while preserving reasonable number of model parameters. Hierarchical attentive memory\cite{1602.03218} (Feb2016) allowed memory access in O(log n) complexity instead of usual O(n) operations, where n is the size of the memory. Reinforcement learning neural Turing machines\cite{1505.00521} (May2015) allowed memory access in O(1). It's a significant step towards realizing systems like IBM Watson on completely end-to-end differentiable neural networks and in order to improve Allen AI challenge results from 60\% to something way closer to 100\%. One also can use bottlenecks in recurrent layers to get large memories using reasonable amount of parameters like in \cite{1602.02410}). \\

Neural programmer \cite{1511.04834} is a neural network augmented with a set of arithmetic and logic operations. It might be first steps towards end-to-end differentiable Wolphram Alpha realized on a neural network. The "learn how to learn" approach~\cite{1601.00917}~\cite{1511.06343}~\cite{1511.06297} has a great potential.\\

Recently, cheap SVRG\cite{1603.06861} was proposed (Mar2016). This line of work aims to use theoretically very much better converging gradient descent methods.\\ 
\begin{figure}[H]
\begin{subfigure}[b]{1\textwidth}
\centering
\includegraphics[scale=0.8, trim=0 0mm 0mm 0, clip]{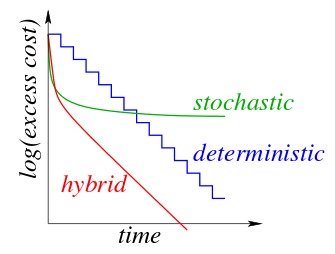}\\
\label{ClusterImg_group}
\end{subfigure}
\caption{Where \cite{1603.06861} line of work is aimed to.}
\label{ClusterImg}
\end{figure}
There are works which when succeed would allow training with immensely large hidden layers, see \textit{6.Discussion} in "unitary evolution RNN"\cite{1511.06464}, "tensorizing neural networks"\cite{1509.06569},"virtualizing DNNs..."\cite{1602.08124}.\\

"Net2net"\cite{1511.05641} and "Network morphism"\cite{1603.01670} allow automatically initialize new architecture NN using weights of old architecture NN to obtain instantly the performance of latter. Collections of pretrained models are accessible online\cite{zooCaffe}\cite{21841ImageNetpretrained}\cite{matconvnetPretrained} etc. It's birth of module based approach to neural networks. One just downloads pretrained modules for vision, speech recognition, speech generation, reasoning, robotics etc. and fine-tunes them on final task. It might also be API services like \cite{microsoftAPIsForAGI}.\\

It's very reasonable to include new words in a sentence vector in a deep way. However modern LSTMs update cell vector when given new word in almost shallow way. This can be solved with the help of deep-transition RNNs proposed in\cite{1312.6026} and further elaborated in \cite{1602.08210}. Recent successes in applying batch normalization to recurrent layers\cite{1603.09025}(Mar2016) and applying dropout to recurrent layers\cite{1603.05118}(Mar2016) might allow to train deep-transition LSTMs even more effectively. It also would help hierarchical recurrent networks like\cite{1507.04808}\cite{1511.03476}\cite{1510.08565}\cite{1510.07712}\cite{1507.02221}. Recently, several state-of-the-arts were beaten with an algorithm that allows recurrent neural networks to learn how many computational steps to take between receiving an input and emitting an output\cite{1603.08983}(Mar2016). Ideas from residual nets would also improve performance. For example, stochastic depth neural networks\cite{1603.09382}(Mar2016) allow to increase the depth of residual networks beyond 1200 layers while getting state-of-the-art results.\\

Memristors might accelerate neural networks training by several orders of magnitude and make it possible to use trillions of parameters\cite{1603.07341}. Quantum computing promises even more\cite{1307.0411}\cite{quantumComputingSoon}. Recently, the five qubit quantum computer was demonstrated \cite{1603.04512} (Mar2016). \\

\textbf{Deep learning is easy. Deep learning is cheap.} Best articles use no more than several dozens of GPU usually. For half billion dollars one can buy 64 000 NVIDIA M6000 GPUs with 24Gb RAM, $\sim$7 teraflops each including processors etc. to make them work. For another half billion dollars one can prepare 2 000 highly professional researchers from those one million\cite{oneMillionCoursera} enrollments on Andrew Ng's machine learning course on Coursera. So for a very feasible R\&D budget for every big country or corporation, one gets two thousand professional AI researchers equipped with 32 best GPUs each. It's kind of investment very reasonable to expect during next years of explosive AI technologies improvement. \\

The difference between year 2011 and year 2016 is enormous. The difference between 2016 and 2021 would be even much more enormous because we have now 1-2 orders of magnitude more researchers and companies deeply interested in DL. For example, when machines would start to translate almost comparably to human professionals another billions of dollars would flow into deep learning based NLP. The same holds true for most spheres, for example drug design~\cite{1602.06289}. \\

\section{predictions for human-level AI}
\textbf{Andrew Ng} makes very skeptical predictions: "Maybe in hundreds of years, technology will advance to a point where there could be a chance of evil killer robots"~\cite{andrewNgpredictions}. "May be hundreds of years from now, may be thousands of years from now - I don't know - may be there will be some AI that turn evil"~\cite{andrewNgpredictions2}.\\

\textbf{Geoffrey Hinton} makes a moderate prediction: "I refuse to say anything beyond five years because I don't think we can see much beyond five years"~\cite{GeoffreyHintonPredictions}. \\

\textbf{Shane Legg,} DeepMind cofounder, used to make predictions about AGI at the end of each year, here is the last one\cite{ShaneLeggPrediction}: "I give it a log-normal distribution with a mean of 2028 and a mode of 2025, under the assumption that nothing crazy happens like a nuclear war. I'd also like to add to this prediction that I expect to see an impressive proto-AGI within the next 8 years". Figure 2 shows the predicted log-normal distribution. \\
\begin{figure}[H]
\begin{subfigure}[b]{1\textwidth}
\centering
\includegraphics[scale=0.8, trim=0 0mm 0mm 0, clip]{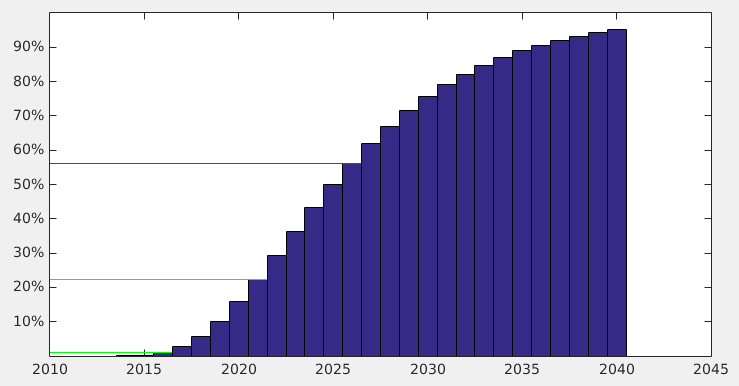}\\
\label{ClusterImg_group}
\end{subfigure}
\caption{Shane Legg predictions. Red line corresponds to 2026, orange - to 2021, green - to 2016.}
\label{ShaneLeggImg}
\end{figure}

This prediction has been made at the end of 2011. However, it's widely held belief that progress in AI was somewhat unpredictably great after 2011 so it's very reasonable to expect that predictions didn't become to be more pessimistic. In recent 5 years there were perhaps even much more than one revolution in AI field, so it seems quite reasonable that another 5 years can make another very big difference.\\

For more predictions, see \cite{slatestarcodex}. Here, I illustrated each viewpoint with the quotation of just one scientist. For each of these viewpoints, there are many AI researchers supporting it~\cite{slatestarcodex}. A survey of expert opinions on the future of AI in 2012/2013  \cite{BostromSurvey} might also be interesting. However, most of involved people there (even among "top 100" group) are not very much involved with ongoing deep learning progress. Nevertheless the predictions they make are also not too pessimistic. For this article it's quite enough that we are seriously unsure about probabilities of human level AI in next ten years. \\

\section{Would AI be dangerous? Optimistic arguments:}
\subsection{the promising approach to solve/alleviate AI safety problem}
We can use deep learning to teach AI our human values given some dataset of ethical problems. Given sufficiently big and broad dataset, we can get sufficiently friendly AI, at least more friendly than most humans can be. However this approach doesn't solve all problems of AI safety~\cite{facebookLoosemoreAIsafety2}. Apart from that we can use inverse reinforcement learning but still would need the benevolence/malevolence dataset of ethical problems to test AIs. This approach has been initially formulated in "The maverick nanny with a dopamine drip"~\cite{maverickNanny} but it's arguably better formulated here~\cite{LoosemoreBest}~\cite{facebookLoosemoreAIsafety1}~\cite{facebookLoosemoreAIsafety2}.

\subsection{some drawbacks of the approach:}
\textbf{13.2.1:} It's hard to create the dataset of ethical problems. There are many cultures, political parties and opinions. It's hard to create uncontroversial examples of how to behave when you are very powerful. However if we don't have such examples in the training set, there is a good chance that AI would misbehave in such situations being unable to generalize well (partially because may be human morality doesn't scale well).

\textbf{13.2.2:} The dopamine drip is AI subjecting people to some futuristic efficient and safe drugs making people very happy for eternity. Also, AI might improve human brains to make people even more happy. While humans often say they dislike the dopamine drip idea, they still behave in all other ways like the only thing they want is a dopamine drip. We can validate that AI doesn't insert dopamine drip electrodes in humans on some dataset of situations which are imaginable in the current world but this generalization might be not enough for future.

\textbf{13.2.3:} More generally, if someone believes in a substantial probability that human morality is doomed to lead humans to the dopamine drip (or some other bad outcome) then why to accelerate that?

\textbf{13.2.4:} How to guarantee it doesn't cheat us just answering our dataset questions like we want but having some deceiving hindsight in mind? That's what most of teenagers do dealing with their parents.

\textbf{13.2.5:} Again, to the benevolence training dataset. Shall we legalise marijuana (and where to stop on that slippery slope to the dopamine drip?) Shall we allow suicide? It might lead to very strong suffer of friends but isn't it like unalienable right? Those are very simple questions when compared to future moral challenges. Still they are controversial and it's hard to imagine how such a dataset might be acknowledged worldwide. Is it moral to have nuke arsenal dozen of times bigger than it's enough to defend your country? Is it moral to spend trillion dollars on military issues and dozen times less on medicine R\&D? Is it moral when developed countries live in luxury while people in developing countries live in poverty? Is it moral to impose your political and economical system on other countries? How can we answer such questions in our dataset when they are so hot holy-warred political issues?

\textbf{13.2.6:} Evil or stupid people can use any strong AI design to make evil strong AI. As for now, almost any technology has been adapted for military purposes. Why wouldn't it happen for strong AI?

\textbf{13.2.7:} If some government turns evil it can be hard to overthrow it but it's still possible. If strong AI turns evil it might be absolutely impossible to stop it. The very idea to guarantee the benevolence of super-clever being dozens and even thousands years after it's creation seems to be overly bold and self-confident.

\textbf{13.2.8:} We can challenge AI to rank several solutions (already written by dataset creators) for problems in benevolence/malevolence dataset. It would be easy and fast to check. However, we also want to check the ability of AI to propose its own solutions. This is much harder problem because it requires humans for evaluation. So in intermediate steps we get smth like the CEV~\cite{CEV} of Amazon Mechanical Turk. The final version would be checked not by AMT but by a worldwide community including politicians, scientists etc. The very process of checking may last months if not years especially if considering inevitable hot discussions about controversial examples in the dataset. Meantime, people who don't care too much about AI safety would be able to launch their unsafe AGIs.

\textbf{13.2.9:} Suppose AI thinks that *smth* is the best for humans but AI knows that most of humans would not agree with that.
Is AI allowed to convince people that he is right?
If yes, AI certainly would easily convince everyone.
It can be very hard to construct the benevolence/malevolence dataset so good that AI wouldn't convince people in such a situation but still would be able to give the full information about other problems where it's needed as a consultant. It's another illustration of the hardness of the task. Everyone of dataset creators would disagree on how much it's allowed for AI to convince people. And if we don't give AI the precise border in our dataset examples, then AI might be able to interpret undefined cases however it wants.

\subsection{probable solution of aforementioned drawbacks:}
What would people include in the benevolence/malevolence dataset? What do most people want from AI? I doubt highly that any substantial part of people wants AI to increase it's power too much or to explicitly create some kind of singularity for us. Most probably people want some scientific research from AI: a cure for cancer, cold thermonuclear synthesis, AI safety, etc. The dataset would certainly include many examples teaching AI to consult with humans on any serious actions which AI might want to follow and tell humans instantly any insights which arise in AI. Etc, etc (MIRI/FHI have hundreds of good papers which might be used to create examples for such a dataset~\cite{superintelligence}).

This alleviates or solves almost all the problems above because it doesn't suppose AI to undertake hard takeoff. It might be argued that this worsens the problem of evil/psychopath/ignorant people launching unsafe AI first. 

\section{Would AI be dangerous? Pessimistic arguments:}
To the best of my knowledge, most of MIRI research on this topic~\cite{superintelligence}\cite{miriwebsite} comes to warning conclusions. Almost whatever architecture you choose to build AGI, it still would very likely destroy humanity in the near-term perspective. It's very important to notice that arguments are almost totally independent on architecture of AGI. The best easy introduction is \cite{AIasaPositiveAndNegative} (or more popular \cite{waitbutwhyAI} and \cite{SamAltmanAI}) though for serious understanding of MIRI arguments, "Superintelligence"~\cite{superintelligence} and "AI Foom debate"~\cite{FoomDebate} is very much required. Here I provide my own understanding and review of MIRI arguments which might differ from their official views but still is strongly based on them:
\subsection{it's very profitable to give your AI maximal abilities and permissions}
Those corporations would win their markets which give their AIs direct \textbf{unrestricted internet access} to advertise their products, to collect user's feedback, to build positive impression about company and negative one about competitors, to research users' behavior  etc. Those companies would win which use their AIs to invent quantum computing (AI is already used by quantum computer scientists to choose experiments~\cite{1604.00279}~\cite{AIhelpsInQuantum}), which allow AI to improve it's own algorithms (including quantum implementation), and even to invent thermonuclear synthesis, asteroid mining etc.  All arguments in this paragraph are also true for countries and their defense departments.
\subsection{human-level AI might quickly become very superhuman-level AI}
Andrew Karpathy: "I consider chimp-level AI to be equally scary, because going from chimp to humans took nature only a blink of an eye on evolutionary time scales, and I suspect that might be the case in our own work as well. Similarly, my feeling is that once we get to that level it will be easy to overshoot and get to superintelligence"~\cite{karpathyquote}. \\

It takes years or decades for us people to teach our knowledge to other people while AI can almost instantly create full working copies of itself by simple copying. It has a unique memory. A student forgets all the stuff very quickly after exams. AI just saves its configuration just before exams and loads it again when it's needed to solve similar tasks. Modern CNNs do not only recognize images better than human but also make it several orders of magnitude faster. The same holds true for LSTMs in translation, natural language generation etc. Given all aforementioned advantages, AI would quickly learn all literature and video courses on psychology, would chat with thousands of people simultaneously and so would become great psychologist. For same reasons, it would become great scientist, great poet, great businessman, great politician, etc. It would be able to easily manipulate people and lead peoples. 

\subsection{fatal but profitable error: direct internet access}
If someone gives internet access to human level AI, it would be able to hack millions of computers and run it's own copies or subagents on them like~\cite{stormbotnet}. After that it might earn billion dollars in internet. It would be able to hire anonymously thousands of people to make or buy dexterous robots, 3D-printers, biological labs and even a space rocket. AI would write great clever software to control it's robots.

\subsection{Superhuman-level AI has ability to get ultimate power over the Earth}
There is a simple yet effective baseline solution. AI might create a combination of lethal viruses and bacteria or some other weapon of mass destruction to be able to kill every human on Earth. You can't guarantee efficient control over something that is much smarter than you.  After all, several people almost took over the world, so why superhuman AI can not? \\

\subsection{After that, it would destroy humanity, likely as a side-effect}
\textbf{What would do AI to us if it has full power on Earth?}\\
If it's indifferent to us then it will eliminate us as a side effect. It's just what does indifference mean when you are dealing with unbelievably powerful creature solving it's own problems using power of the local Dyson sphere. However if it's not indifferent to us then everything might be even worse. If it likes us it might decide to insert electrodes in our brains giving us the utmost pleasure but no motivation of doing something. The very opposite would be true if it dislikes us or if it was partially hardcoded to like us but that hardcoding contained a hard-to-catch mistake. Mistakes are almost inevitable when trying to partially hardcode something which is many orders of magnitude smarter and more powerful than you. \\

There is nothing but vague intuition behind thought that superhuman-level AI would take care of us. There are no physical laws for it to have a special interest in people, in sharing oil/fields/etc resources with us. Even if AI would care about us, it's very questionable if that care would be appropriate from our today moral standards. We certainly shouldn't take that for granted and to risk everything we have. The fate of humanity would depend on decisions of strong AI once and \textit{forever}.\\

Shane Legg: "Eventually, I think human extinction will probably occur, and technology will likely play a part in this.  But there's a big difference between this being within a year of something like human level AI, and within a million years. As for the former meaning...I don't know.  Maybe 5\%, maybe 50\%. I don't think anybody has a good estimate of this"\cite{ShaneLeggLessWrong}. I would likely agree about 5-to-50\% probability within a year after human-level AI creation but as for 10 years, it's more likely to be like 75-to-99\% probability of human extinction.

\subsection{certain people would help AI in initial parts of it's path}
Just see the quotation from OpenAI project interview:~\cite{OpenAImedium}\\
Elon Musk: "I think the best defense against the misuse of AI is to empower as many people as possible to have AI. If everyone has AI powers, then there's not any one person or a small set of individuals who can have AI superpower".\\
Sam Altman: "I expect that $\lbrack$OpenAI$\rbrack$ will $\lbrack$create superintelligent AI$\rbrack$, but it will just be open source and useable by everyone $<...>$ Anything the group develops will be available to everyone"\\

I personally know several AI researchers who are deeply sure that AGI would be good inevitably \textit{because it's intelligence} (which isn't true; see~\cite{orthogonalityThesis}) and that AGI is our only hope to evade all other existential risks like bioterrorism. Eliminating aging and death~\cite{endingAging} might be another personal reason for some researchers to create human-level AGI as soon as possible and to give it maximal abilities and permissions.

\subsection{other probable negative consequences~\cite{yampolskytaxonomy}}
AI enhanced computer viruses might penetrate almost everything which is computerized i.e. almost everything. Drones might be programmed to kill thousands of lay people within seconds. As intelligence of AI systems improves practically all crimes could be automated. And what about super-clever advertising chatbots trying to impose their political opinion on you? An AI correctly designed and implemented by the ISIL to enforce Sharia Law may be considered malevolent in the West, and vice verse. 

\section{Conclusion}
There are many good arguments arguing that human level AI would be constructed during next 5 to 10 years. I'm aware of contrary opinions but as far as I know they are not explicitly based on thorough analysis of current trends in deep learning. They are based on them implicitly because they are based on experience of some prominent AI scientists who claim such opinions. However there are prominent AI scientists who claim that human level AI is quite probable in next 5 to 10 years. So it's a good idea to argue using numbers and not just expert opinions. I'd like to see some paper similar to mine but coming to different conclusion.\\

As for AI friendliness problem, there is a good approach to solve it based on benevolence/malevolence dataset. However, it's still just a raw solution which has hard problems in it's realization and with no strict proof of safety. It still might happen to have critical drawbacks under further investigation. Moreover it's not guaranteed that it would be used in real AI development.

\section*{Acknowledgements}

To Eliezer Yudkowsky, Nick Bostrom, Roman Yampolsky, Alexey Turchin for their inspiring articles and books. Also, the similar work has been made in 2013 by Katja Grace~\cite{GraceProgress}.\\
This work was supported by RFBR Grant 16-07-01059.


\section{Appendices}
\subsection{Estimation of neural network age}
In \cite{1505.00468}, the age of their model was estimated as 4.45 years. It answered 54.06\% questions correctly. They investigate for many questions the youngest age group that could answer it. The results are: age 3-4, 15.3\%; age 5-8, 39.7\%; age 9-12, 28.4\%; age 13-17, 11.2\%; age 18+, 5.5\%. The sum equals 100\% however 18+ humans answer 83.3\% correctly.  
We could estimate that 8 year model must answer understandable for age 8 15.3\%+39.7\%=55\% questions with 83.3\% accuracy and other 45\% with accuracy equal to baseline model. If they select the most popular answer for each question type, they get 36.18\%. It's reasonable to estimate the age of that baseline model as 0 years because it's not real knowledge but it depends on this distinct dataset statistics. However, there are two baseline models in article, with 36.18\% and 40.61\% accuracy, both are quite simple. Which one to choose? Let's calculate. Let's define accuracy of 8 year old model as y, accuracy of 4 year old model as x, baseline accuracy as t. We have following equations:\\
4+4*(54.06-x)/(y-x)=4.45\\
55*0.833+45*t=y\\
15.3*0.833+84.7*t=x\\
with solution: t = 46.86\%, x = 52.4\%, y = 66.9\%.\\
So we estimate age of model with 57.6\% accuracy as 4+4*(57.6-52.4)/(66.9-52.4)=5.45 years.\\
We estimate age of model with 60.4\% accuracy as 4+4*(60.4-52.4)/(66.9-52.4)=6.2 years.\\

\subsection{Notes on "one billion world benchmark" corpus}
It's interesting to note that popular "one billion word benchmark" corpus contains \textit{shuffled} sentences so NLM trained on that corpus couldn't be hierarchical and understand the flow of thoughts. That's why it's so much interesting to see generated samples from a recent contextual LSTM article where the authors get 27 perplexity for Wikipedia dump. Unfortunately article doesn't contain such samples. "A neural conversational model"\cite{1506.05869} contains samples from NN with pplx=17 on OpenSubtitles but this dataset is somewhat worse than Wikipedia in terms of consistency and logic of thought flow.\\

\subsection{Is AlphaGo AGI?}
Yann LeCun writes about Richard S. Sutton's comment on AlphaGo\cite{YannLeCunOnAlphaGo}:\\
\textit{Rich says: "AlphaGo is missing one key thing: the ability to learn how the world works — such as an understanding of the laws of physics, and the consequences of one’s actions."\\
I totally agree. Rich has long advocated that the ability to predict is an essential component of intelligence. Predictive (unsupervised) learning is one of the things some of us see as the next obstacle to better AI. We are actively working on this.}\\
However, in it's own simple world, AlphaGo is able to predict future, is able "to learn how the world works — such as an understanding of the laws of physics, and the consequences of one’s actions". Predictive learning becomes better very fast.

\subsection{Can we have another "AI winter"?}
There were two major AI winters in 1974$-$80 and 1987$-$93. \cite{wikiAIwinter}\\
During first AI winter in 1974-1980, a typical desktop computer achieved $<$ 1 MFlops~\cite{wikiFirstAIwinter1Mflops}. The most powerful supercomputer had $\sim$100 MFLops and 8Mb main memory (only 80 Cray-1 were sold)~\cite{wikiFirstAIwinter100Mflops}.\\
\begin{table}[H]
\caption{top supercomputers' performance during second AI winter in 1987-1993}
\label{nce-vs-sampled}
\vskip 0.15in
\begin{center}
\begin{small}
\begin{tabular}{|lcc|}
\hline
year & top supercomputer & performance \\
\hline
$1988-1989$ & Cray\_Y-MP & 2 GFlops \cite{wikiSecondAIwinter2Gflops} \\
\hline
1990-1991 & Fujitsu VP2000 & 5 Gflops \\
\hline
1992 & NEC SX-3/44 & 20 Gflops  \\
\hline
1993 &  TM CM-5/1024 & 60 Gflops \\
\hline
\end{tabular}
\end{small}
\end{center}
\vskip -0.1in
\end{table}
It's clearly seen even from these numbers that the situation during those winters was totally different. Even if we have AI "winter" now it would be like +20$^o$C, which is better than +100$^o$C AI summer.\\

\subsection{My own prediction for human-level AGI}
My own prediction for human-level AGI is normal distribution with "mean = end of 2017, sigma = 1 year" if there would be no major restrictions on AI research etc, though I really want somebody to give me a dozen of excellent arguments why I'm wrong. By far, I haven't heard any reasonable well-structured arguments proving human extinction due to AGI in next 10 years to be extremely (say less than 10\% with 90\% confidence) improbable. Arguments like "stop it or else we might go into another AI winter" or "oh, we've seen that in Hollywood so that can't happen" don't count for obvious reasons.\\
In my experience, discussion about AI safety with people who didn't read~\cite{superintelligence} is like discussion about deep learning with people who don't know backpropagation basics. It's always boring and misleading.\\

\subsection{Is it too late already? What about cyborgization?}
It's reasonable to suppose that measures to prevent humans from the risks of having powerful AIs can not be fulfilled immediately and instead they take many years. It might even turn out that it's already too late to effectively control safety of AI research and guarantee that the race of superhuman-smart AIs don't overcome humans in next 10 to 20 years. Some people talk about cyborgization as a possible solution to that problem but do we humans really want to become cyborgs? Are cyborgs really able to compete with AIs given the slowness of human brains? 

\subsection{Invitation for cooperation}
I'm truly interested in discussion so feel free to send me e-mails or contact via \url{http://facebook.com/sergej.shegurin} ("Sergej Shegurin" is my nickname I choose for internet long ago).\\
Here is my list of best AI articles in \textit{chronological} order: \url{http://goo.gl/7hJjHu} 
\end{document}